# A high-resolution nationwide urban village mapping product for 342 Chinese cities based on foundation models

Lubin Bai[1], Sheng Xiao[2], Ziyu Yin[2], Haoyu Wang[2], Siyang Wu[3], Xiuyuan Zhang[2], Shihong Du[2]

[1]Institute of Remote Sensing and GIS, Peking University, Beijing 100871, China
[2]College of Urban and Environmental Sciences, Peking University, Beijing 100871, China
[3]Environment & Society, Faculty of Science School of Earth, Mcmaster University, Hamilton, Ontario, L8S 4K1, Canada

*Correspondence to*: Shihong Du (shdu@pku.edu.com)

**Abstract.** Urban Villages (UVs) represent a distinctive form of high-density informal settlement embedded within China's rapidly urbanizing cities. Accurate identification of UVs is critical for urban governance, renewal, and sustainable development. But due to the pronounced heterogeneity and diversity of UVs across China's vast territory, a consistent and reliable nationwide dataset has been lacking. In this work, we present GeoLink-UV, a high-resolution nationwide UV mapping product that clearly delineates the locations and boundaries of UVs in 342 Chinese cities. The dataset is derived from multisource geospatial data, including optical remote sensing images and geo-vector data, and is generated through a foundation model-driven mapping framework designed to address the generalization issues and improve the product quality. A geographically stratified accuracy assessment based on independent samples from 28 cities confirms the reliability and scientific credibility of the nationwide dataset across heterogeneous urban contexts. Based on this nationwide product, we reveal substantial interregional disparities in UV prevalence and spatial configuration. On average, UV areas account for 8 % of built-up land, with marked clustering in central and south China. Building-level analysis further confirms a consistent low-rise, high-density development pattern of UVs nationwide, while highlighting regionally differentiated morphological characteristics. The GeoLink-UV dataset provides an open and systematically validated geospatial foundation for urban studies, informal settlement monitoring, and evidence-based urban renewal planning, and contributes directly to large-scale assessments aligned with Sustainable Development Goal 11. The GeoLink-UV dataset introduced in this article is freely available at https://doi.org/10.5281/zenodo.18688062 (Bai et al., 2026).

## 1 Introduction

Urbanization in China over the past several decades represents one of the most significant human-environment transformations in modern history. Within this rapid process, a unique kind of residential area known as "Urban Villages" (UVs), has emerged as a distinct feature of the Chinese urban landscape (Liu et al., 2010; Cao et al., 2025; Wang, 2022). It can be considered as a distinctive type of urban informal settlement within the Chinese context, but unlike other informal settlements such as slums formed via squatting in many Global South cities (Un-Habitat, 2004), UVs are original rural settlements that have been spatially encompassed by expanding urban territories while retaining their collective land ownership status due to dual urban-rural land system of China (Chan and Wei, 2021; Chung, 2010; Wu, 2012). Morphologically, UV regions are characterized by a dense aggregation of low-rise buildings and narrow roads, often appearing as distinct islands within the formal urban fabric (Wang et al., 2025b; Lin et al., 2024) (Fig. 1). Functionally, they generally serve a dual role: while often criticized for lagging infrastructure and overcrowding, they provide essential, affordable housing for millions of migrant workers, acting as a crucial landing pad for new urban populations (Hao et al., 2012; Kochan, 2015).

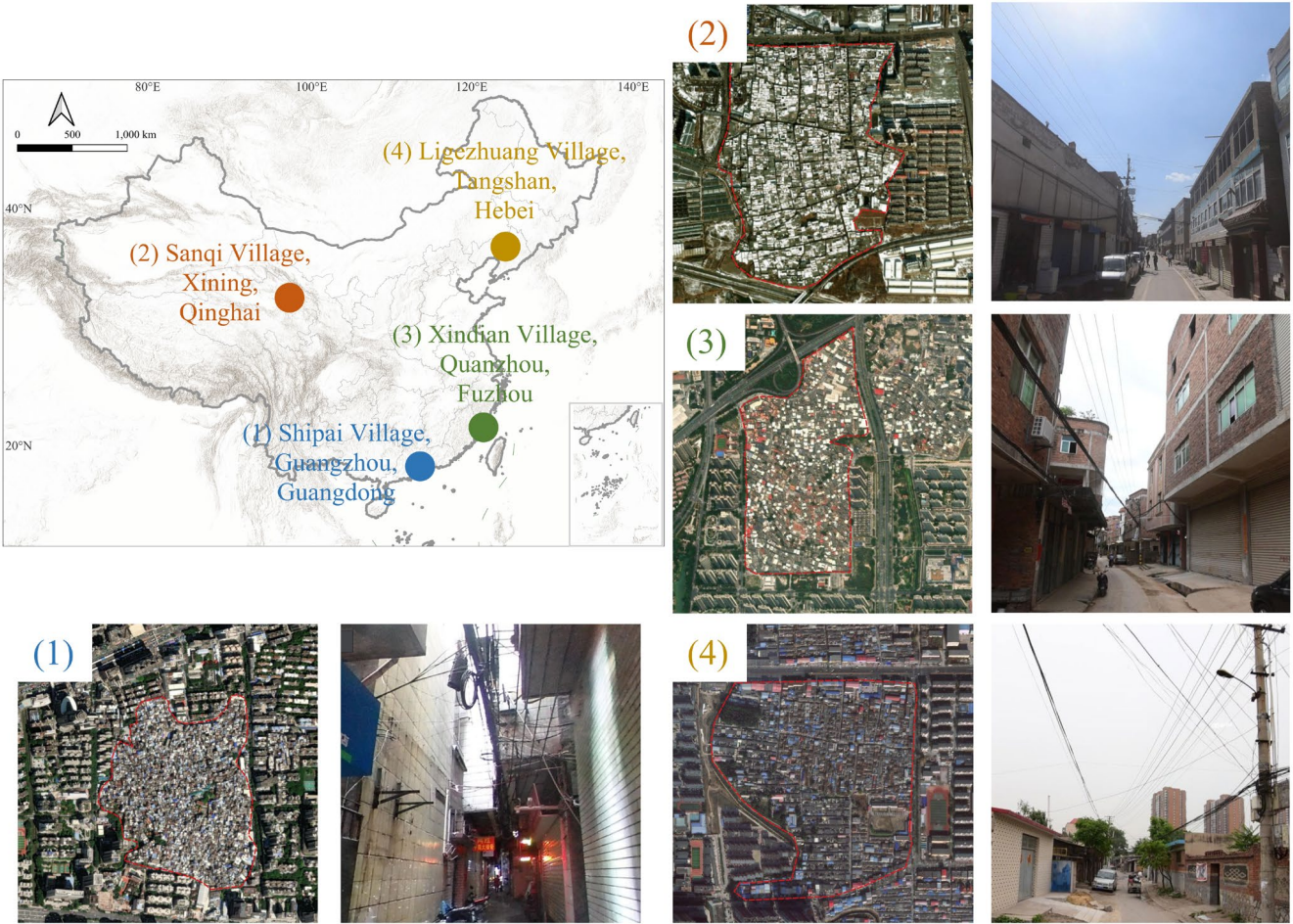

**Figure 1:** Urban villages in RS and street-view images (Sources: Esri and Baidu Map | Powered by Esri and Baidu Map).

The accurate mapping and monitoring of UVs are significant for urban sustainable development. As the Chinese government shifts the focus toward high-quality urbanization and urban renewal, the redevelopment of UVs has become a policy priority (China, 2025). Furthermore, from a global perspective, mapping UVs contributes directly to the United Nations' Sustainable Development Goal 11 (SDG 11), which calls for making cities inclusive, safe, resilient, and sustainable (Colglazier, 2015;

Sachs, 2022). However, despite their significance, there remains a critical shortage of available and reliable mapping product on UVs, particularly at a national scale (Cao et al., 2025). Traditional acquisition methods relying on field surveys are labor-intensive, time-consuming, and incapable of keeping pace with the large-scale dynamic evolution of urban footprints. And there is a growing trend toward employing remote sensing (RS) image and machine learning techniques for the automated identification of UVs (Zhang et al., 2025; Zhang et al., 2024b).

The capacity of RS images to deliver large-scale, high-resolution data on physical morphology (e.g., building density, layout, and texture) enables the identification of UVs by capturing their distinct visual characteristics within the urban landscape. Early attempts predominantly adopt classification-based approaches, which involve delineating spatial units (e.g., parcels) and subsequently extracting and aggregating RS features for identification (Huang et al., 2015; D'oleire-Oltmanns et al., 2011). However, the performances of these methods heavily depend on the rationality of the spatial division, and the complex feature aggregation process hinders the implementation of efficient end-to-end learning. Consequently, a growing number of research has shifted towards modeling UV identification as a semantic segmentation task, aiming to directly delineate boundaries from RS image (Wei et al., 2023; Zhang et al., 2025; Chen et al., 2019). This pixel-level paradigm, which facilitates end-to-end learning, has progressively become the mainstream methodology. Nevertheless, UVs often exhibit spectral similarities with other urban features (e.g., contiguous industrial plants), leading to "visual confusion" that hinders accurate identification based solely on RS image. To mitigate this, recent studies have integrated ground-level geospatial data, such as Street-view image (Chen et al., 2022a; Fan et al., 2022) and Points of Interest (POIs) (Chen et al., 2022b). These multi-source data provide complementary functional semantic information, significantly enhancing the accuracy and reliability of UV identification.

Despite these advancements, current research is predominantly confined to specific cities, struggling to generalize to nationwide scales (Cao et al., 2025). The primary bottleneck lies in the profound spatial heterogeneity of UVs across China's diverse geographic regions (Fig. 1), resulting in the limited scalability and transferability of most existing models. To address the generalization challenges, recent efforts have explored advanced learning strategies, such as semi-supervised learning (Fan et al., 2025; Lin et al., 2024) and Mixture-of-Experts (Lee et al., 2025). Notably, Lin et al. (2024) employ a curriculum labeling strategy to iteratively annotate representative labels, achieving nationwide, long-time-series UV mapping. However, this work lacks a comprehensive large-scale accuracy assessment, leaving the reliability of the resulting data product uncertain. Consequently, there is an urgent need to develop UV identification models with robust generalization capabilities to facilitate nationwide UV mapping. Furthermore, systematic evaluation is essential to ensure the reliability of such data products, thereby meeting the rigorous demands of urban planning and governance.

Fortunately, the recent emergence of Foundation Models (FMs) presents a transformative paradigm shift that offers new solutions to these generalization and scalability bottlenecks (Zhu et al., 2026; Bai et al., 2025). Unlike traditional deep learning models that are typically trained on limited, task-specific datasets, FMs are pre-trained on massive data with vast parameters, endowing them with powerful representation learning capabilities and generalization potential(Deepseek-Ai et al., 2025; Zhang et al., 2024a; Gui et al., 2024). For example, Segment Anything Model (SAM) can segment objects accurately with appropriate prompts (Kirillov et al., 2023), and it have been validated in small-scale UV identification task (Zhang et al., 2024b). In the geospatial domain, RS FMs have been specifically tailored to accommodate the unique characteristics of RS image (Xiong et al., 2024; Hong et al., 2024; Wang et al., 2025a; Xiao et al., 2024; Bai et al., 2025), and been used in some downstream tasks, like burn scar mapping (Szwarcman et al., 2024). However, studies that utilize geospatial FMs for large-scale and systematical mapping, especially in complex urban scenarios, remain scarce. There is a critical need to investigate how these models can be effectively integrated into mapping workflows (e.g., sample collection, large-scale prediction, and result refinement) to empower robust, nationwide-scale analysis.

To bridge this gap, this study proposes GeoLink-UV, a novel multimodal framework leveraging the power of FMs to achieve the high-resolution, nationwide mapping of UVs across 342 Chinese cities. This dataset clearly delineates the locations and boundaries of UVs in 342 Chinese cities for circa 2023. A geographically stratified accuracy assessment based on independent samples from 28 cities yields an overall F1-score of 0.77 and an Intersection over Union (IoU) of 0.60, confirming the reliability and scientific credibility of the nationwide dataset across heterogeneous urban contexts. By defining UVs through a morphological and functional lens rather than restrictive administrative criteria, this dataset offers a highly selective resource that allows users to customize data layers according to specific research or governance needs. Leveraging this national-scale product, we conduct comprehensive analyses of UV patterns from the macro to the micro scale, including national distributions, intra-urban spatial configurations, and building-level characteristics, yielding policy-relevant insights to inform urban renewal and governance. Methodologically, this work also contributes a systematic and transferable paradigm for applying FMs to robust, large-scale urban mapping tasks.

The remainder of this paper is organized as follows: Sect. 2 describes the source data and methods used for developing and evaluating the urban village data. Sect. 3 presents the description and evaluation results of the GeoLink-UV data product. Sect. 4 analyzes the UV spatial characteristics in China and discusses the UV definition. Sect .5 summarizes the conclusions and discusses the limitations of the study. The final section details the data acquisition methods.

## 2 Materials and Methods

Achieving nationwide UV mapping hinges on the generalization capability of the overall framework. In this study, we address this challenge by leveraging FMs, exploiting their powerful pre-trained representation capabilities to acquire high-quality training samples and design a robust mapping architecture applicable across diverse urban landscapes. As illustrated in Fig. 2, we propose GeoLink-UV, a comprehensive mapping framework consisting of four sequential stages: heuristic sample collection, GeoLink-UV training, SAM refinement, and accuracy assessment. The key components of this framework are detailed below.

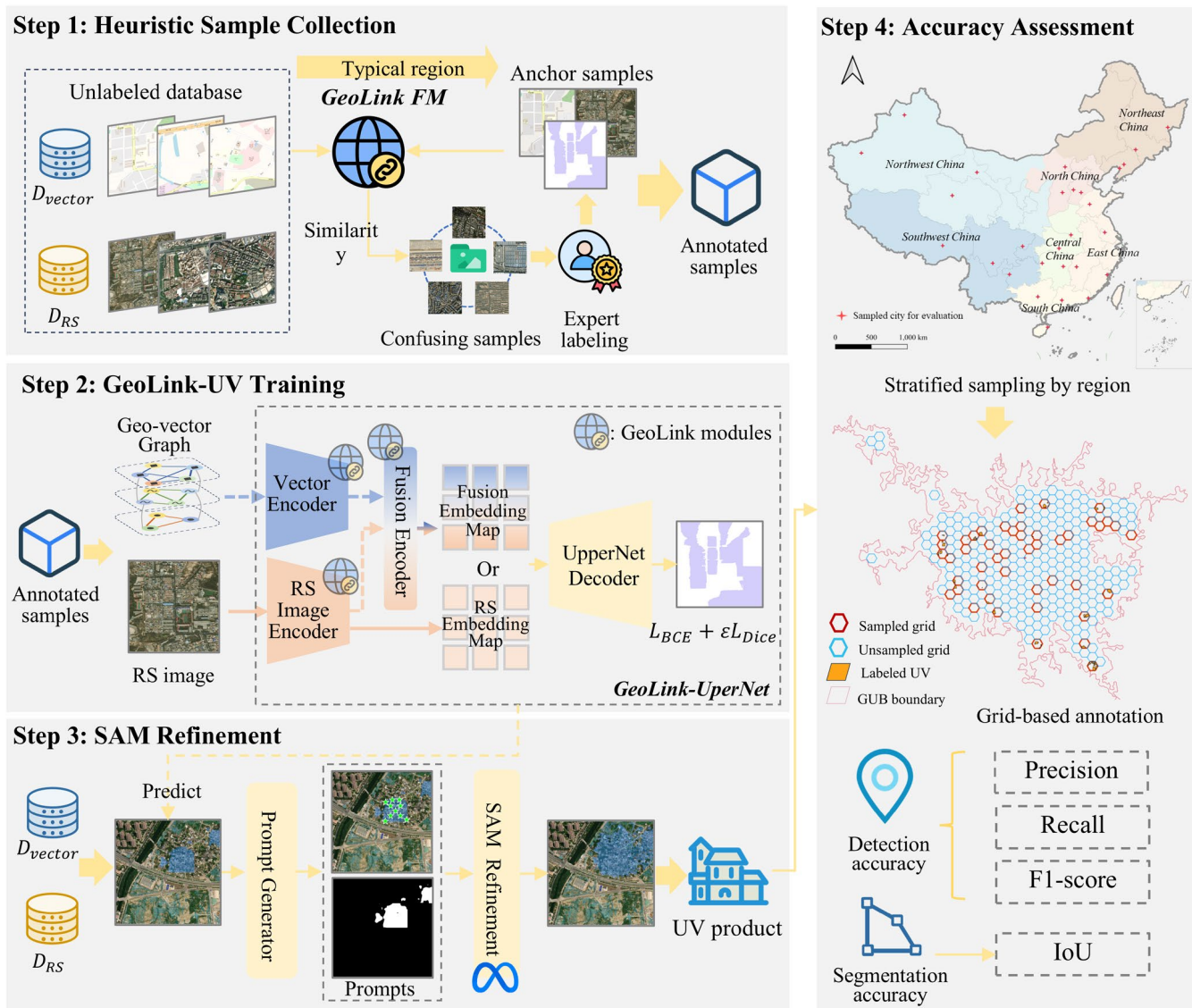

**Figure 2:** The workflow of GeoLink-UV (Sources: Esri and OpenStreetMap | Powered by Esri and OpenStreetMap).

## 2.1 Introduction of GeoLink and SAM

To ensure generalization and mitigate issues related to regional heterogeneity, like feature confusion and boundary inaccuracy, the proposed framework is built upon two FMs: GeoLink, a multimodal geospatial FM, and SAM, a prompt-driven universal segmentation model.

GeoLink jointly encodes optical RS images and geo-vector data (e.g., OSM), enabling patch-entity level multimodal fusion ideal for fine-grained tasks like semantic segmentation. It contains three encoders: a RS encoder, a geo-vector encoder, and a fusion encoder. First, the RS encoder is a vanilla ViT-L, which divides the image into patches and processes them as a sequence.

It generates patch-level RS embeddings $\mathcal{E}_R \in \mathbb{R}^{L_P \times D_P}$, where $L_P$ is the number of RS image patches and $D_P$ is the patch feature dimension, i.e. $D_P$=1024 following the default set of ViT-L. Second, the geo-vector encoder is built upon heterogeneous graph neural network, and can process three types of geometries, including point, polyline, and polygon. It utilizes the language model CLIP-Bert (Devlin et al., 2018; Radford et al., 2021) to encode the semantic attributes of geo-vectors, thus enabling an open-vocabulary encoding scheme that can dynamically adapt to and incorporate unrestricted semantic attributes of geo-vectors. It can output the embeddings for each individual geo-vector entity $\mathcal{E}_V \in \mathbb{R}^{L_V \times D_V}$, where $L_V$ is the number of geo-vector entities and $D_V$ is the feature dimension, Third, the fusion encoder is a two-way attention based Transformer that takes $\mathcal{E}_R$ and $\mathcal{E}_V$ as inputs, and generate patch-level multi-modal fusion embeddings $\mathcal{E}_F \in \mathbb{R}^{L_P \times D_F}$. It can be seen that the spatial shape $L_P$ of $\mathcal{E}_F$ keeps consistent with the feature map $\mathcal{E}_R$ produced by the RS encoder, thus allowing them to be directly fed into downstream network modules (e.g., decoder heads) with identical structures for task-specific processing. The fusion encoder integrates physical attributes from RS images with ground-level semantic details from geo-vectors, and can construct comprehensive urban scene understanding. As shown in Fig. 2, GeoLink is utilized in both the sample collection and UV identification phases in this study.

SAM is a promptable FM for general image segmentation, capable of generating high-quality object masks from various interactive prompts such as points, boxes, or masks. Notably, trained on an extensive high-quality dataset, SAM exhibits remarkable zero-shot generalization, allowing it to accurately delineate object boundaries when provided with appropriate prompts. This capability has led to its frequent integration into segmentation frameworks, where upstream models provide prompts to guide SAM toward precise segmentation. For instance, SAM has been coupled with object detection models for instance segmentation (Ren et al., 2024; Diab et al., 2025) or used to refine semantic segmentation outputs (Lin et al.; Kweon and Yoon, 2024). In this study, we leverage SAM to assist in the refinement of UV boundaries during the inference and nationwide mapping phase.

## 2.2 Data Sources

This study utilizes two primary data types: RS images and geo-vector data. For RS images, considering large-scale accessibility, we utilize Esri World Imagery at a $1-meter$ spatial resolution with RGB bands (https://www.arcgis.com/home/item.html?id=10df2279f9684e4a9f6a7f08febac2a9, last access: 14 August 2025). Esri World Imagery integrates data from multiple sources to offer seamless, cloud-optimized coverage with sub-meter to meter-level resolution. The service is continuously updated, and we utilize the images closest to 2023 in this study. As for the geo-vector data, they are primarily sourced from OSM (https://wiki.openstreetmap.org/, last access: 14 August 2025). To address potential OSM data gaps in China, Points of Interest (POIs) data from Amap Service (https://lbs.amap.com/, last access: 24 December 2023) are supplemented, and they are combined with OSM data to construct the input geo-vector dataset. Notably, GeoLink supports open-category processing, enabling it to seamlessly integrate and process geo-vector data from diverse sources. For more preprocessing details of geo-vector data, please refer to (Bai et al.). After collecting the original RS images and geo-vector data, we project all of them into the same geographic coordinate system for spatial alignment. In addition, we employ the Global Urban Boundary (GUB) data (Li et al., 2020) to define the urban built-up extent for each city, thereby extracting UVs located within these defined built-up areas.

## 2.3 Methods

### 2.3.1 Heuristic Sample Collection

High-quality, representative samples are essential for training a model capable of nationwide generalization. In this section, we describe our heuristic sample collection strategy, designed to efficiently curate a high-quality training set. Prior to annotation, a clear operational definition of UVs is required. As aforementioned, the term "Urban Village" represents a unique form of informal settlement emerging from the distinct institutional and rapid urbanization context of China. In this study, to enable large-scale mapping, we adopt an operational definition grounded in both morphological and functional characteristics, which allows us to distinguish urban villages from the formal urban fabric using RS images and geo-vector data. Drawing on prior research (Lin et al., 2024; Cao et al., 2025), this study scopes the definition of UV as follows: (1) **Morphology:** continuous built-up areas with high building density, typically consisting of low-rise, small-scale structures and irregular, narrow road networks; (2) **Function:** spaces characterized by functional hybridity, primarily residential, but often incorporating commercial as well as historical or cultural uses.

Based on previous studies and considering the geographic diversity (Cao et al., 2025; Lin et al., 2024; Wang et al., 2025b; Zhang et al., 2025), 12 cities representing distinct topographic and climatic zones are selected for sample annotation: Beijing, Zhengzhou, Changsha, Chengdu, Chongqing, Guiyang, Guangzhou, Shanghai, Shenzhen, Wuhan, Xi'an, and Xining. To facilitate the selection of representative samples within cities and improve the efficiency of sample collection, we propose the heuristic sample collection strategy (see Fig. 2), which utilizes region similarity to guide annotation through the following steps:

- **Initial Annotation:** For a given city, a small set of typical UVs (e.g., Shipai Village in Guangzhou) are identified based on news reports and literature to form an initial sample set $S = \{s_1, s_2, ..., s_n\}$. These serve as anchors for subsequent expansion.
- **Similarity Calculation:** Leveraging GeoLink's representation power, we calculate regional similarity to guide sample screening. The city was divided into $512m \times 512m$ grids. For any unlabeled grid $g_u$, its feature similarity with the initial sample set is calculated as:

$$\alpha_{sim} = \frac{1}{n}\sum_{i=1}^{n} \frac{Mean(\mathcal{E}_{g_u})^T Mean(\mathcal{E}_{g_{s_i}})}{\|Mean(\mathcal{E}_{g_u})\|\|Mean(\mathcal{E}_{g_{s_i}})\|} \tag{1}$$

where $g_{s_i}$ is the grid containing the anchor sample $s_i$, $\mathcal{E}_{g_{s_i}}$ and $\mathcal{E}_{g_u}$ is the output embedding of GeoLink for grid $g_{s_i}$ and $g_u$, and $Mean(\cdot)$ denotes the average along the patch dimension. A higher $\alpha_{sim}$ indicates a higher likelihood of confusion with UV characteristics.

- **Confusing Sample Annotation:** We observe that incorporating negative samples (confusing non-UV areas) significantly reduces misclassification. Consequently, similarity scores are ranked, and we check through the top 5% and find non-UV areas with high confusion potential to annotates as negative samples. Since precise boundary delineation is not required for negative samples, this process is highly efficient. Additionally, UV areas with relatively low similarity scores (e.g., ranked in 10% − 30%) to anchors (indicating heterogeneity) are also annotated to increase diversity.

Five professionals with backgrounds in urban planning and RS perform the annotation, referencing street views and commercial maps to ensure reliability. Following this heuristic collection, images are cropped to 256 × 256 pixels, yielding a final dataset of 5470 positive UV samples and 6916 negative samples. It is also worth noting that this workflow is training-free, as we directly utilize the pre-trained GeoLink model for feature extraction.

### 2.3.2 Model Training

As illustrated in Fig. 2, we design a UV segmentation network named GeoLink-UperNet. This network accepts RS images and constructed geo-vector graph as inputs, utilizing GeoLink's three encoders as the backbone for feature extraction and fusion to obtain patch-level multi-modal fusion embeddings $\mathcal{E}_F$. For the decoder, we employ the Unified Perceptual Parsing Network (UperNet) (Xiao et al., 2018). UperNet is based on a Feature Pyramid Network (FPN), enabling the hierarchical fusion of multi-scale features while capturing both rich spatial details and high-level semantic context. Its concise yet powerful architecture makes it highly effective for complex semantic segmentation tasks in urban scenarios (Deng et al., 2024; Bolcek et al., 2025). In this framework, the UperNet decoder process the GeoLink's fusion features $\mathcal{E}_F$ and output UV segmentation results. To optimize GeoLink-UperNet, we combine binary cross-entropy loss $L_{BCE}$ and Dice loss $L_{Dice}$ (Milletari et al., 2016). The former measures the pixel-wise distribution difference between predictions and ground truth, ensuring classification confidence, while $L_{Dice}$ directly optimizes the overlap between predicted and actual regions. They can be calculated as follows:

$$L_{BCE} = -\frac{1}{N}\sum_{i=1}^{N}[y_i \cdot \log(P_i) + (1 - y_i) \cdot \log(1 - P_i)] \quad (2)$$

$$L_{Dice} = 1 - \frac{2\sum_{i=1}^{N} y_i P_i + \mu}{\sum_{i=1}^{N} y_i + \sum_{i=1}^{N} P_i + \mu} \quad (3)$$

where $N$ is the number of pixels, $y_i$ and $P_i$ are the label and predicted results of $i-th$ pixel, and $\mu = 1e - 7$ is the smoothing term. Finally, the combined loss function is defined as $L = L_{BCE} + \varepsilon L_{Dice}$, where $\varepsilon$ is set to 0.01 to balance boundary clarity with overall shape sensitivity.

### 2.3.3 SAM Refinement and Nationwide Mapping

Preliminary UV predictions for all cities are generated using the trained GeoLink-UperNet. To further refine segmentation boundaries during the mapping phase, we employ SAM for post-processing. As a promptable segmentation model, the input prompts play a crucial role in SAM because these prompts provide localization guidance of intended objects. As shown in Fig. 2, we design a prompt generator to convert the UV mask output by GeoLink-UperNet into a set of spatial prompts to guide SAM for pixel-level refinement. Given the initial binary segmentation mask, the prompt generator first conducts morphological preprocessing to remove small fragments and thin inter-region bridges, ensuring that each remaining region corresponds to a meaningful local structure. Then, we collaborate two kinds of prompts, i.e., point and mask, to generate refined results.

- **Point.** Determining the most salient prompt points is challenging. We adopted a simple yet empirically effective strategy: (1) sampling the geometric centroid of a given mask instance, as the center of an object tends to be positive and feature-discriminative; (2) compressing region contours to a controlled number of representative vertices via Ramer-Douglas-Peucker simplification, and then generating inward-outward offset points as additional positive/negative cues for SAM.
- **Mask.** Based on our experience and prior studies (Lin et al.; Dai et al., 2023; Zhang et al., 2023), mask prompts impose strong constraints, often leading to outputs highly consistent with the input mask. Leveraging this, we utilize the mask as an auxiliary prompt only when the point-prompt-based SAM output meets specific criteria: (1) the maximum confidence score of SAM outputs is below 0.6, or (2) the IoU between the refined result and the original mask is below 0.7. These thresholds are empirically tuned.

Unlike the small patch input ($256 \times 256$ pixels) used for GeoLink-UperNet, we clip RS images at a resolution of $1024 \times 1024$ for SAM input to provide richer context and eliminate tiling artifacts. Finally, this refinement process is applied to GeoLink-UperNet predictions for every city, resulting in a precise, nationwide UV map.

**2.4 Accuracy Assessment**

Taking geographic heterogeneity into account, we devise a rigorous protocol to select assessment samples for accuracy assessment of the mapping results at a national scale. As illustrated in Fig. 2, the assessment samples are annotated in the following steps:

- **Stratified sampling by region:** Excluding the 12 cities utilized for training, we select 28 test cities based on China's seven major geographic divisions, including Northeast, North, Central,, East, South, Southwest and Northwest. As shown in Fig. 2, four cities of varying sizes are sampled from each division as the assessment region to ensure representativeness.
- **Annotation strategy:** To construct the ground truth, each test city is tessellated into hexagonal grid units. We then randomly sample and manually annotate 15% of these grids for evaluation purposes, as the example shown in Shijiazhuang in Fig. 2. Ultimately, a total of 40.6 $km^2$ of UV regions across the 28 cities are labeled for evaluation.

Crucially, this evaluation protocol ensures that the test set is spatially disjoint from the training set and covers a broad geographic range, thereby effectively verifying the model's out-of-distribution performance and generalization capability for national-scale UV mapping. For all predicted UV boundaries, we evaluate both detection and segmentation accuracy. Detections are considered true positives if they overlap with annotated areas, enabling the calculation of precision, recall, and F1-Score and segmentation performance is measured using the IoU.

**2.5 Implementation Details**

Several key aspects of the framework implementation warrant clarification: (1) Missing geo-vector data and single-modal fallback: Geo-vector coverage is incomplete in some regions, which can prevent GeoLink-UperNet from performing multimodal inference., and we design an adaptive fallback strategy. Because GeoLink's multimodal embeddings have the same spatial layout as its RS encoder outputs (see Section 2.1), we implement a straightforward single-modal fallback: by removing the vector and fusion encoders and connecting the RS encoder directly to the UperNet decoder, we train an RS-only UV segmentation model with the same decoder head. During inference, the multimodal pipeline is used whenever geo-vector data are available; otherwise the RS-only model is used. (2) Training hyperparameters: The GeoLink-UperNet is optimized using the Adam optimizer with weight decay of 0.01. We set the learning rates to $3e-6$ for GeoLink encoders and $3e-4$ for the decoder. The model is trained for 40 epochs with a batch size of 32. (3) Train/validation split: From the annotated positive samples we randomly hold out 20% as a cross-validation set. The remaining positive samples together with all negative samples form the training set. Furthermore, to rigorously evaluate the final mapping product, we annotate independent test samples across 28 cities, the details of which are provided in Section 3.2. (4) SAM Variant: For the refinement module, we employ SAM-HQ (Ke et al., 2023) with a ViT-H backbone. This variant improves upon the original SAM architecture by significantly enhancing boundary precision and the overall quality of segmentation masks.

## 3 Results

### 3.1 Description of the GeoLink-UV product

We map the locations and coverage of UVs across 342 cities in mainland China using the GeoLink-UV framework. To reveal national spatial patterns, we calculate the proportion of UV area to the total built-up area (defined by the GUB) for each city. These distributions are visualized via a graduated symbol map and an inset frequency histogram in Fig. 3a. First, the frequency histogram indicates that over 65% of cities exhibit UV proportions below 10%, with a national average of 8%. This suggests a substantial improvement in urban infrastructure following decades of modernization and urbanization. Notably, Shangri-La City emerges as the sole outlier with a proportion exceeding 30%. As a world-renowned tourist destination, Shangri-La's urban core retains a substantial number of historical residential structures, which fall within our definition of UVs and contribute to this exceptionally high ratio. Second, a pronounced spatial heterogeneity is observable across the country in the graduated symbol map. Clusters of cities with high UV proportions, with some exceeding 18.2%, are concentrated in central-northern (specifically Henan and Hebei provinces) and southeastern (Guangdong and Fujian provinces) parts of China. In contrast, cities in northeastern region, particularly in Heilongjiang, generally display significantly lower UV proportions, typically below 5.3%. This disparity likely stems from the Northeast's history as an early industrial base, where extensive urban redevelopment occurred during the planned economy era, leading to the earlier clearance of UVs compared to other regions.

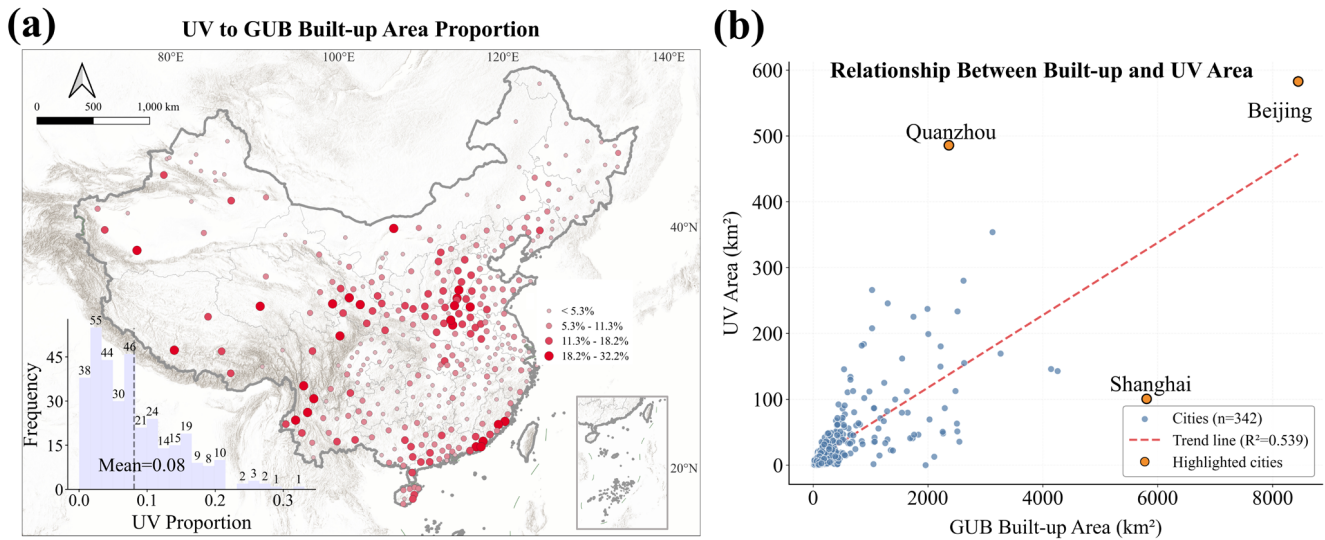

**Figure 3:** Mapping results and statistical analysis of UVs in China derived from GeoLink-UV.

To investigate the relationship between UV presence and urban scale, we perform a linear regression analysis with GUB built-up area as the independent variable and UV area as the dependent variable (Fig. 3b). First, we can observe that a large number of points are clustered in the lower-left corner of the scatter plot, indicating that the UV area in most cities is less than 100 square kilometers. Second, the results reveal a relative strong positive correlation with $r^2 = 0.539$. Residual analysis identifies Shanghai and Quanzhou as the most significant deviations below and above the trend line, respectively. This divergence implies that, beyond mere city size, region-specific development and governance policies are critical determinants of UV scale. Beijing, comparable in size to Shanghai, aligns closely with the regression trend, highlighting distinct urbanization trajectories even among megacities.

**3.2 Accuracy Assessment**

As shown in Fig. 4, we evaluate the nationwide mapping results across seven regions using the annotated assessment samples from 28 cities described in Section 2.5. GeoLink-UV achieves an overall mapping accuracy of 0.77 in F1-Score and 0.60 in IoU across the nation, indicating its robust capability in identifying both the locations and spatial extents of UVs in diverse regions. Regionally, Central China attains the highest precision, indicating fewer false positives, while East China achieves the highest recall, reflecting strong detection completeness but with relatively lower precision. Northeast China records the highest F1-Score, suggesting a well-balanced performance. In contrast, North China shows comparatively lower F1 due to reduced precision. For segmentation accuracy, Northeast China achieves the highest IoU, whereas Southwest China records the lowest IoU. Despite competitive detection performance in the Southwest, boundary delineation remains challenging, likely due to visual ambiguity between UVs and surrounding residential buildings, as further discussed in Section 3.3. Moreover, the evaluation is based on manually annotated samples from 28 cities spanning diverse geographic and socio-economic contexts, ensuring representative coverage across different settlement patterns. The consistent performance across most regions further demonstrates the model stability and generalization capability. These results collectively confirm the reliability and robustness of the nationwide mapping product.

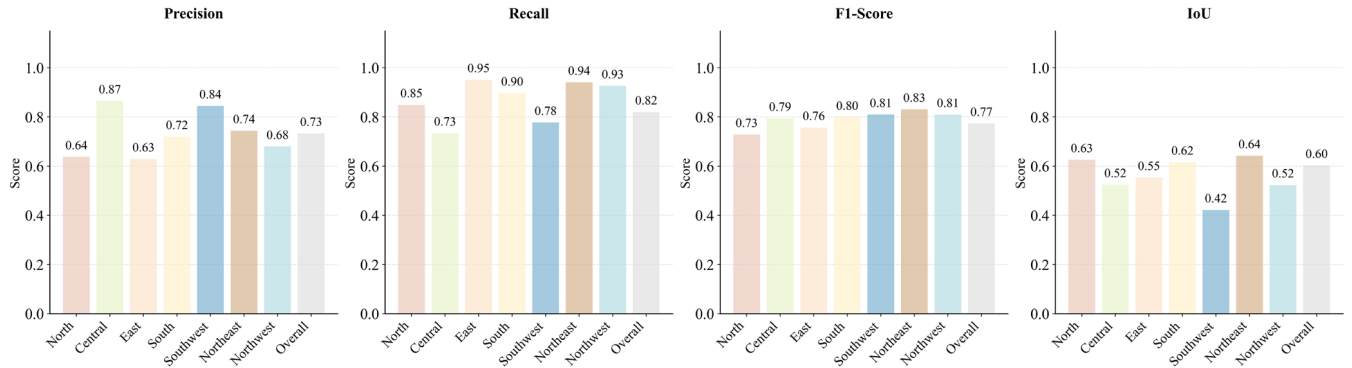

**Figure 4:** Accuracy assessment of GeoLink-UV in seven regions.

We further visualize the mapping details of GeoLink-UV in Fig. 5 across diverse cities. First, GeoLink-UV is capable of accurately detecting UVs with diverse sizes, which provides the foundation for its high recall performance. Regardless of the size or spatial extent of the UV areas, the model can effectively identify them without missing fragmented or relatively small clusters. Second, the detailed results demonstrate that GeoLink-UV achieves highly precise boundary delineation. It clearly outlines the shapes and extents of UVs and effectively distinguishes them from surrounding roads and formal residential or commercial buildings in most cases. This strong segmentation performance can be largely attributed to the integration of multimodal data, which introduces richer semantic information, as well as the SAM-based post-processing module that refines and optimizes boundary accuracy. Nevertheless, some local inaccuracies remain in the dataset. In particular, low-rise buildings adjacent to UVs, especially industrial plants with similar morphological characteristics, are occasionally misclassified and incorporated into the mapped UV boundaries. Finally, the five example cities are located in different regions of China and differ substantially in terms of natural conditions and socioeconomic contexts, leading to considerable morphological variations in UVs. Nevertheless, GeoLink-UV maintains consistently high prediction performance across these heterogeneous settings, indicating that the resulting data product possesses strong nationwide generalizability and reliability.

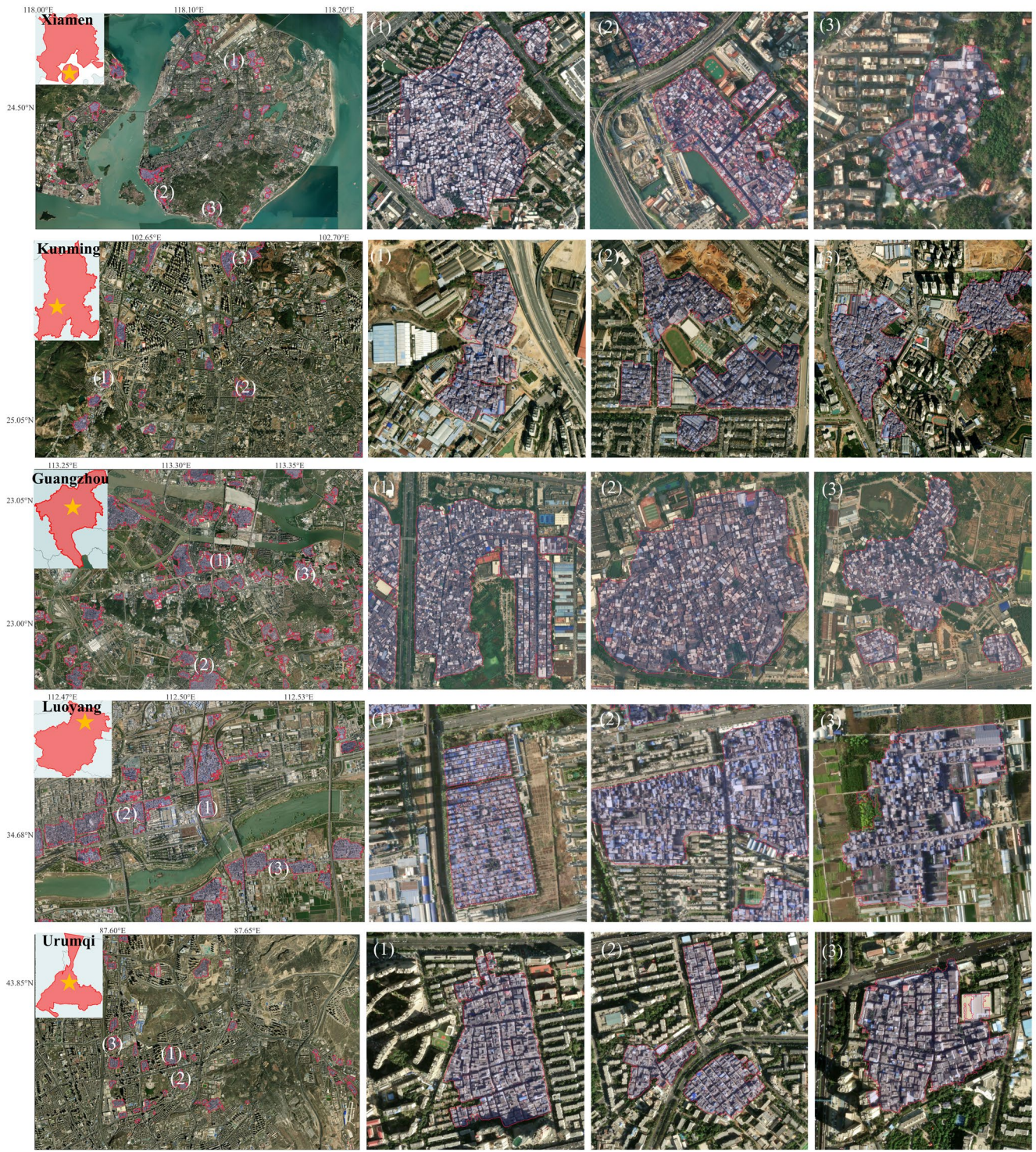

**Figure 5:** Mapping details of GeoLink-UV (Sources: Esri | Powered by Esri).

**3.3 Comparative Assessment with Existing Products**

**Table 1. Evaluation results of three mapping products**

| Metric   | Urban-Scene | LtCUV | GeoLink-UV |
|----------|-------------|-------|------------|
| F1-Score | 0.690       | 0.743 | 0.798      |
| IoU      | 0.539       | 0.584 | 0.621      |

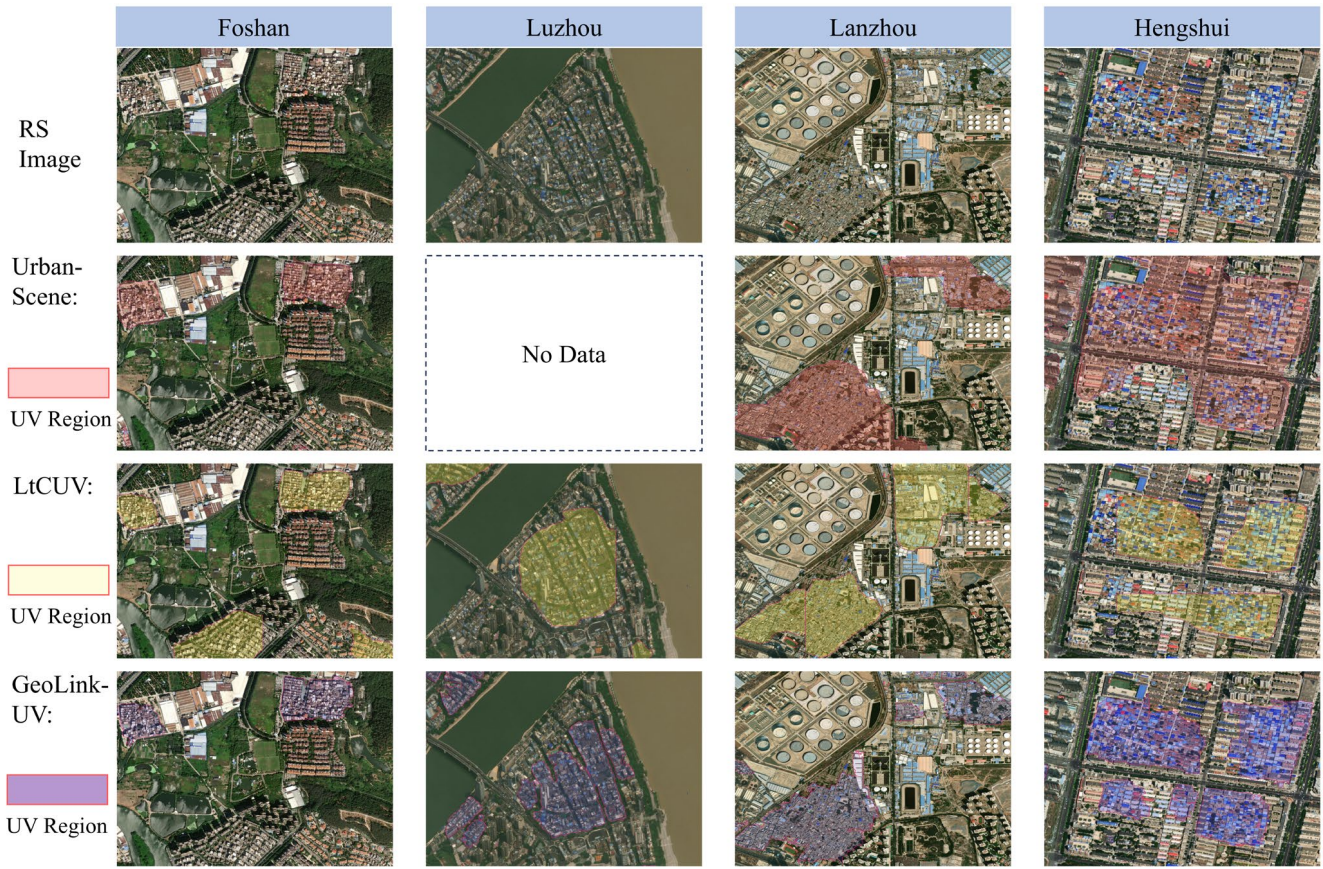

**Figure 6:** Comparison with existing UV products in detail (Sources: Esri | Powered by Esri).

To benchmark the performance of GeoLink-UV mapping results, we compare it against two established UV products: (1) Urban-Scene (Du et al., 2020), a functional zone mapping product covering 81 major Chinese cities that includes a specific UV category; and (2) LtCUV (Lin et al., 2024), a multi-temporal UV product, from which we use the 2023 edition to align with our study period. To facilitate a fair comparison given the varying spatial coverages of these products, validation samples were rigorously selected from the overlapping regions within the 28 test cities. As summarized in Table 1, GeoLink-UV outperforms both existing products.

To provide a granular assessment of these differences, we visualize mapping results from four representative regions in Fig. 6. First, the RS images (row 1) highlight substantial morphological heterogeneity across different regions. Notably, in Luzhou (Southwest China), UVs exhibit high visual ambiguity with surrounding features, significantly complicating accurate extraction. This morphological confusion is the primary factor contributing to the relatively lower IoU observed in the Southwest region. Second, GeoLink-UV demonstrates a significantly lower misclassification rate compared to the other two products. As evidenced in the Foshan case, our model effectively distinguishes low-density villas from UVs despite their shared low-rise characteristics. Similarly, in Lanzhou, it successfully differentiates industrial plants, which characterized by large-area single structures, from UV clusters, a distinction that competing products frequently fail to make. Finally, the results in Hengshui reveal that GeoLink-UV produces superior boundary delineation. The model accurately traces UV extents along natural and artificial boundaries, such as rivers and major roads, aligning closely with human visual interpretation standards. We attribute this boundary precision to the robust segmentation capabilities inherent in the SAM architecture used by GeoLink-UV.

### 3.4 Ablation Studies

Table 2. Evaluation results of ablation studies.

| Ablation | Setting | Precision | Recall | F1-Score | IoU |
|---|---|---|---|---|---|
| Default | / | 0.733 | 0.820 | 0.774 | 0.603 |
| Segmentation model | DeepLabV3+ | 0.688 | 0.653 | 0.670 | 0.526 |
|  | Segmenter | 0.701 | 0.664 | 0.682 | 0.531 |
| Multimodal effect | GeoLink-Uni | 0.703 | 0.801 | 0.749 | 0.589 |
|  | Scale-Mae | 0.692 | 0.734 | 0.712 | 0.555 |
| SAM effect | Without SAM | 0.718 | 0.827 | 0.769 | 0.584 |

To systematically evaluate the contribution of individual components within the GeoLink-UV framework, we conduct a series of ablation studies, with quantitative results summarized in Table 2. The analysis focuses on three key aspects.

**Segmentation model.** We replace the semantic segmentation model with two representative traditional architectures, i.e., DeepLabV3+ (Chen et al., 2018) and Segmenter (Strudel et al., 2021), to benchmark performance against GeoLink-UV. Both alternatives exhibit significant performance degradation compared to GeoLink-UV. We attribute this decline primarily to the absence of extensive pretraining on massive RS datasets in these traditional models, which severely limits their generalization capabilities for large-scale geographic mapping tasks.

**Multimodal effect.** We evaluate two variants relying solely on RS images for inference: GeoLink-Uni and Scale-Mae (Reed et al., 2023). GeoLink-Uni, which retains the multi-modal pretraining of our framework but utilizes only RS images for inference, shows a noticeable performance degradation. Specifically, compared with other metrics, detection precision drops more significantly. This suggests that without complementary ground-level information from VGI during inference, the model is prone to misclassifying urban regions that bear a morphological resemblance to UVs, leading to increased false positives. Furthermore, Scale-Mae, a FM pretrained exclusively on RS images, underperforms GeoLink-Uni, underscoring the critical value of multi-modal strategies in the pretraining phase for robust feature representation.

**SAM effect.** Finally, we assess the role of SAM as a boundary refinement module. The exclusion of SAM results in a discernible decrease in IoU, confirming its crucial role in enhancing segmentation accuracy and ensuring precise boundary delineation. Additionally, we can observe that the changes in precision and recall due to the removement of small fragments prior to SAM refinement.

Based on the above results, we can attribute the superior performance of GeoLink-UV to two synergistic factors: first, the inherent power of the underlying FM; and second, the multi-modal architecture of GeoLink-UV. By integrating VGI, the model successfully compensates for the lack of ground-level details in RS image, i.e., granular information that proves critical for the precise discrimination of complex UV regions.

## 4 Discussion

### 4.1 Intra-urban Spatial Patterns of UVs

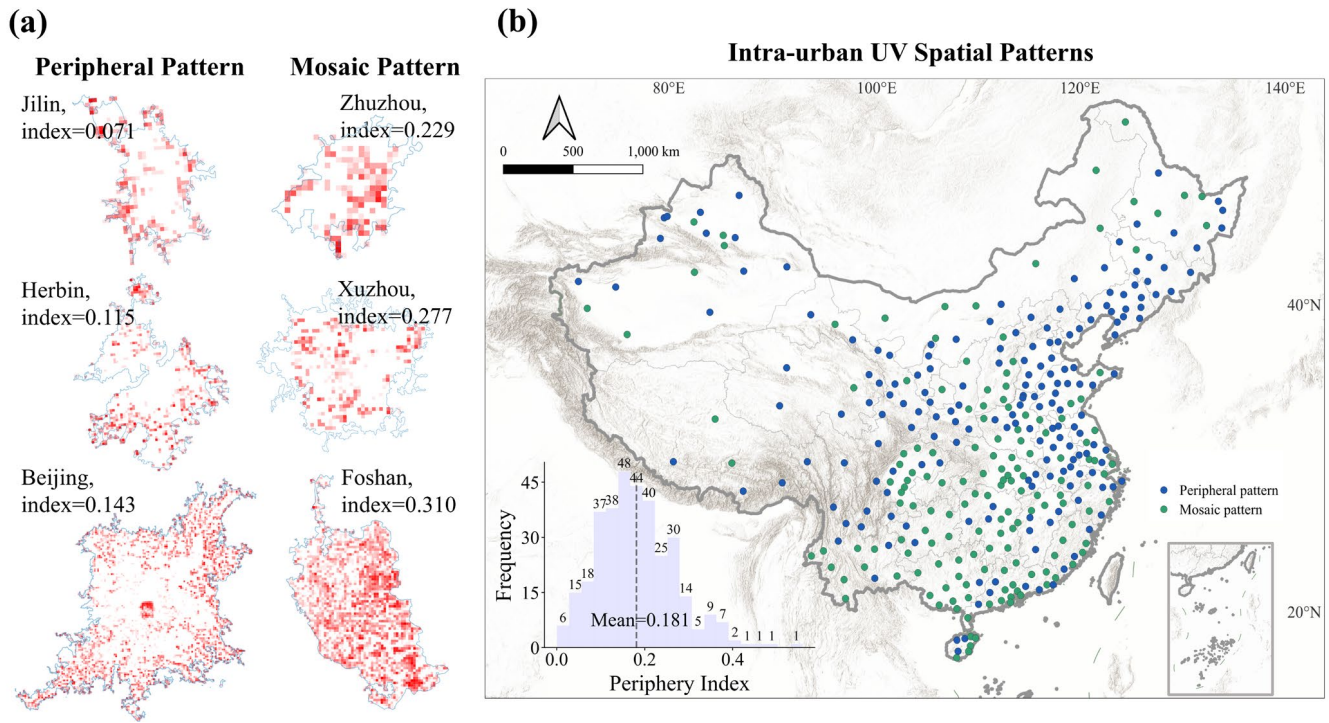

**Figure 7:** Statistical analysis and visualization of intra-urban spatial patterns of UVs.

While the aggregate scale of UVs provides insight into urbanization intensity, it fails to capture the topological relationship between these informal settlements and the formal urban fabric. Understanding where UVs are located and aggregated within a city is critical for policy-making. For instance, UVs encircling a city often act as constraints on urban expansion, whereas UVs embedded deeply within the city primarily pose challenges for internal regeneration and social integration. Therefore, beyond mere area proportions, we further investigate the intra-urban spatial patterns of UVs to reveal their morphological configurations.

To quantify these patterns, we develop a novel metric termed the Periphery Index. Traditional methods often rely on the distance to a city center; however, defining city centers is difficult due to the irregular shapes and polycentric nature of modern cities. Conversely, the urban boundary can be well-defined by the GUB data employed by this study. Consequently, we design and adopt a boundary-referenced approach. The calculation involves three steps: (1) **Distance to the boundary**: For each UV region, we calculate the Euclidean distance from its geometric centroid to the nearest edge of the GUB. (2) **Normalization**: To ensure comparability across cities of varying sizes, this distance is normalized by the city's "Pole of Inaccessibility" (PIA, the point within the GUB that is furthest from any boundary). This yields a value between 0 and 1. (3) **Aggregation**: The final Periphery Index for a city is the area-weighted sum of these normalized values for all its UVs.

The effectiveness of the Peripheral Index in characterizing UV spatial distribution patterns is visually corroborated in Fig. 6a. We select several cities with varying peripheral index values and visualize the distribution of their UVs using grid-based statistics. From the results we can see that with the index increasing, the spatial transition from "encirclement" to "interspersion". Cities with lower index (e.g., Jilin, and Harbin,) exhibit a morphology that UVs are primarily distributed within the urban fringe. Conversely, as the index increases (e.g., Zhuzhou and Foshan), the distribution shifts significantly. In these cities, UVs are not confined to the margin but are deeply embedded within the urban core, interlocking with formal functional zones to form a fragmented, "mosaic-like" texture. The mean peripheral index is computed for the entire set of 342 cities. Each city is subsequently classified as exhibiting either a Peripheral Pattern (index below the mean) or a Mosaic Pattern (index above the mean). Fig. 7b presents the spatial patterns of UVs across all studied cities. We observe a higher frequency of the Peripheral Pattern in the north and a greater proportion of the Mosaic Pattern in the south. This spatial disparity may be attributed to a combination of factors, including regional topography and industrial structure, which warrant further investigation in future research.

## 4.2 Quantified Building-level Characteristic of UV Areas

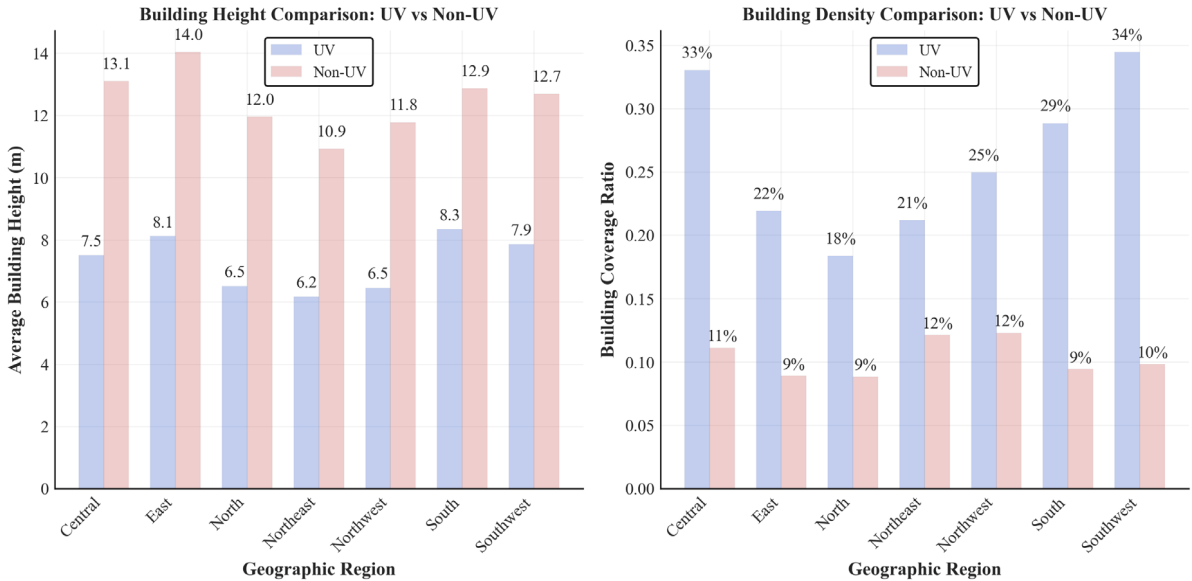

**Figure 8:** Quantified building-level characteristics in UV and non-UV areas.

While the high density and low-rise nature of UVs are widely recognized as defining morphological characteristics, few studies have yet provided quantified, building-level comparative analysis contrasting their architectural attributes against those of formal urban areas (Non-UVs). In this section, combining the GeoLink-UV mapping product and the building footprints and heights dataset from (Papini et al., 2025), we further quantify building-level characteristics of UV and non-UV areas across the seven geographic regions of China. As shown in Fig. 8, two key indicators, including the average building height and building coverage ratio (BCR), are analyzed to reveal the structural distinctiveness of UVs and their regional variations. The findings provide a data-driven basis for differentiated urban renewal policies.

First, the analysis of mean building height reveals a clear pattern of vertical inequality and regional specialization. Across all regions, the average height of UVs is significantly lower than that of Non-UV areas, with height ratios ranging from a low of 0.54 (North) to a high of 0.65 (South). The highest UV mean heights are found in South (8.3 m) and East (8.1 m) China, confirming the intensified vertical expansion of UVs in economically dynamic and land-scarce coastal regions. The largest absolute height differences are observed in East (5.9 m) and Central (5.6 m) China. This signifies the most pronounced vertical segregation in these regions, where formal urban expansion achieves high verticality while UVs lag behind. Second, BCR analysis confirms the extreme horizontal crowding within UVs, a key driver of poor living conditions. UV BCR values range from 18.4% to 34.5%, consistently exceeding those of non-UV areas by 9.1% to 24.6%. The BCR Ratio between UV and Non-UV BCR area exceeds 3.0 in the South (3.05) and Southwest (3.51) regions. The high building density in UVs poses significant challenges regarding ventilation, light access, and emergency response.

Overall, these results quantitatively confirm that Chinese UVs follow a low-rise, high-density building paradigm that is remarkably consistent nationwide, yet modulated by regional socioeconomic and environmental conditions. The regional variations suggest that "one-size-fits-all" redevelopment approaches may be inappropriate. In Southwest China, where density contrasts are most extreme, redevelopment might prioritize density reduction and open space creation. In Northeast China, where morphological differences are less pronounced, incremental upgrading rather than wholesale redevelopment may be more feasible.

**4.3 Definition Scope of UVs**

In this study, we provide a definition of UVs that extends beyond the traditional political or household registration framework, shifting instead toward a morphological and functional perspective. We define UVs as urban residential areas characterized by deficient living conditions, high building density, and functional informalities. This conceptual shift not only addresses the limitations of available geospatial data but also enhances the multi-faceted utility of the resulting dataset.

First, from a policy perspective, UVs are often defined according to the urban-rural dual system, with UVs emerging from rural areas being enveloped by urban boundaries. However, there is almost no publicly accessible geospatial data that includes household registration information. Consequently, we utilize the physical and functional attributes of settlements as an entry point. UVs typically exhibit a low-rise, high-density architectural pattern, irregular and narrow road networks, and a lack of public infrastructure. These characteristics are effectively captured by high-resolution RS imagery and geo-vector data. As outlined in UN-Habitat's framework for informal settlements (Un-Habitat, 2025), a standardized definition of UVs is not strictly necessary for scientific research, provided that the sample sizes are sufficiently large and representative. This allows for generalizable insights and adaptable applications across different regions.

Second, the use of GUB data to define urban areas also means that the final UV extractions are constrained by the geographic limits of these boundaries. While the method is broadly applicable, the number and location of identified UVs may vary depending on the scope of these urban delineations.

Finally, the primary advantage of the morphological-functional definition is its high selectivity and adaptability for diverse user needs. This allows researchers and policymakers to perform customized overlay and filtering operations. For example, users interested in strictly administrative definitions may overlay GeoLink-UV with village boundary maps or land ownership data; those focused on public health outcomes may combine it with household survey data; and those studying informal land markets may integrate it with tenure security indicators. This modularity ensures that GeoLink-UV can support a wide spectrum of applications, from evidence-based urban renewal planning to international comparative research, while respecting the diversity of local definitions and policy needs.

## 5 Conclusions

In this study, we address the critical limitations in identifying UVs in China at national scale by proposing GeoLink-UV, and generate the UV mapping product across 342 cities. By strategically integrating the superior feature generalization of the multimodal GeoLink with the pixel-level segmentation accuracy of SAM, our framework effectively overcomes the challenges posed by UVs vast heterogeneity.  The result is validated for high accuracy and proven reliability across diverse geographic regions. Beyond the product contribution, this work also establishes a robust, innovative paradigm for applying the generalization power of FMs to complex, large-scale geographic mapping tasks. Leveraging the national UV dataset, we reveal consistent yet regionally differentiated patterns in UV spatial distribution and building-level characteristics, offering quantitative insights to inform context-sensitive urban renewal strategies.

Despite the strong performance of GeoLink-UV, several limitations and future directions merit discussion. First, the framework relies on the availability and quality of geo-vector data for multimodal inference. Although a unimodal fallback strategy is employed in regions with missing vector data, the absence or inconsistency of geo-vector information may still affect model performance in data-sparse areas. Second, the current analysis primarily focuses on the physical and morphological characteristics of UVs, and integrating mobility data, street-level, and other human-activity-related datasets could enable a more comprehensive characterization of UV functionality, such as population dynamics, accessibility, and land-use intensity, which are not fully captured by physical form alone.

## 6 Data availability

The GeoLink-UV Dataset for 342 Chinese cities is distributed under the CC BY 4.0 License. The data can be downloaded from the data repository Zenodo at https://doi.org/10.5281/zenodo.18688062 (Bai et al., 2026).

## 7 Code availability

Code used in the research is available upon request.

## Author contributions

Lubin Bai: Writing – review & editing, Writing – original draft, Visualization, Validation, Methodology, Investigation, Formal analysis, Conceptualization. Sheng Xiao: Writing – review & editing, Validation, Resources, Methodology, Conceptualization. Ziyu Yin: Writing – review & editing, Validation, Resources, Methodology, Conceptualization. Haoyu Wang: Writing – review & editing, Writing – original draft. Siyang Wu: Writing – review & editing, Writing – original draft, Validation. Xiuyuan Zhang: Writing – review & editing, Writing – original draft, Methodology. Shihong Du: Writing – review & editing, Resources.

**Competing interests**

The contact author has declared that none of the authors has any competing interests.

**Disclaimer**

Copernicus Publications remains neutral with regard to jurisdictional claims made in the text, published maps, institutional affiliations, or any other geographical representation in this paper. While Copernicus Publications makes every effort to include appropriate place names, the final responsibility lies with the authors. Views expressed in the text are those of the authors and do not necessarily reflect the views of the publisher.

**Acknowledgements**

The work presented in this paper is funded the National Natural Science Foundation of China (No. 42330103) and the National Key Research and Development Program of China (No. 2023YFC3804802). The authors are grateful to all the data contributors who made it possible to complete this research.

**Financial support**

The work presented in this paper is funded the National Natural Science Foundation of China (No. 42330103) and the National Key Research and Development Program of China (No. 2023YFC3804802).